%% file: preprint.tex
\documentclass[11pt,a4paper]{article}
\usepackage{times,latexsym}
\usepackage{url}
\usepackage[T1]{fontenc}
\usepackage[acceptedWithA]{tacl2021v1}

\usepackage[utf8]{inputenc}

\usepackage{microtype}

\usepackage{inconsolata}

\usepackage{times}
\usepackage{latexsym}
\usepackage{lineno}
\usepackage{microtype}
\usepackage{inconsolata}
\usepackage{graphicx}
\usepackage{booktabs}
\usepackage[table]{xcolor}
\usepackage{tcolorbox}
\usepackage{xcolor}
\usepackage{multirow}
\usepackage{amsmath}
\usepackage{amssymb}
\usepackage{hyperref}
\usepackage{url}
\usepackage{tabularx}

\definecolor{lightgray}{gray}{0.9}
\definecolor{darkblue}{RGB}{0, 51, 102}
\tcbuselibrary{skins, breakable}
\newtcolorbox{examplebox}[1]{%
  enhanced,
  colback=lightgray!20,
  colframe=darkblue,
  coltitle=white,
  fonttitle=\bfseries,
  title=#1,
  boxrule=0.5mm,
  sharp corners,
  breakable,
  before upper={\small\obeylines\setlength{\parindent}{0pt}\setlength{\parskip}{0pt}},
  after upper={\par}
}

\newtcolorbox[auto counter]{observationbox}[2][]{%
  enhanced,
  colback=lightgray!10,
  colframe=darkblue,
  coltitle=white,
  fonttitle=\bfseries,
  title={Observation \thetcbcounter: #2},
  boxrule=0.2mm,
  sharp corners,
  breakable,
  fonttitle=\normalsize\linespread{\baselinestretch}\selectfont,
  fontupper=\normalsize\linespread{\baselinestretch}\selectfont,  
  top=0.5ex,                %
  bottom=1mm,             %
  left=2mm,               %
  right=2mm,              %
  toptitle=0mm,           %
  bottomtitle=0mm,        %
  #1
}

\title{On the Utility and Factual Reliability of\\Pruned Mixture-of-Experts Models in the Biomedical Domain}
\author{
\textbf{Atsuki Yamaguchi}{\rm \textsuperscript{1,2}}
\quad
\textbf{Szymon Palucha}{\rm \textsuperscript{2}}
\quad
\textbf{Léo Bijar}{\rm \textsuperscript{2}}
\\
\textbf{Aline Villavicencio}\textsuperscript{1,3,4}
\quad
\textbf{Nikolaos Aletras}\textsuperscript{1}
\\
\textsuperscript{1}University of Sheffield, United Kingdom
\quad\textsuperscript{2}AstraZeneca\\
\textsuperscript{3}University of Exeter, United Kingdom
\quad\textsuperscript{4}Federal University of Rio Grande do Norte, Brazil\\
\texttt{\{ayamaguchi1,a.villavicencio,n.aletras\}@sheffield.ac.uk} 
}

\begin{document}

\maketitle
\begin{abstract}
Mixture-of-Experts (MoE) models offer inference speedups via selective activation but impose substantial memory requirements because the whole network must remain loaded.
Structured expert pruning is a practical approach for reducing deployment costs in resource-constrained settings.
However, prior studies primarily evaluate benchmark utility, leaving the effect of pruning on factual reliability underexplored, particularly in high-stakes domains such as biomedicine.
In this paper, we investigate how domain-specific expert pruning affects both utility and reliability.
We assess four MoE models, six pruning methods, and multiple pruning ratios across generation and classification tasks under in-domain (biomedical) and cross-domain settings.
Results reveal that moderate pruning preserves in-domain utility without immediate reliability decline, although hallucination risks increase at extreme pruning ratios.
When shifting to the general domain, both utility and reliability degrade rapidly.
These findings indicate that safe compression depends heavily on the task and domain. Evaluating pruned MoE models solely on utility is inadequate for high-stakes deployment without reliability assessment.\footnote{Our code is available at \url{https://github.com/gucci-j/moe-pruning-reliability}.}
\end{abstract}

\section{Introduction}
Modern large language models (LLMs) such as Qwen3.6~\citep{qwen36_35b_a3b}, GPT-OSS~\citep{openai2025gptoss120bgptoss20bmodel}, and Nemotron3~\citep{nvidia2025nvidianemotron3efficient} follow a Mixture-of-Experts (MoE) architecture. During the forward pass, MoE models activate only a subset of the network by routing each token through specialized subnetworks called experts~\citep{pmlr-v162-du22c,10937907}. They achieve strong performance while offering substantial inference speedups.
However, this efficiency does not eliminate the memory overhead during deployment.
At inference, all experts must remain loaded in memory, which results in substantially larger memory footprints than those of dense models with comparable active parameter counts.

\begin{figure}[t]
    \centering
    \includegraphics[width=0.95\linewidth]{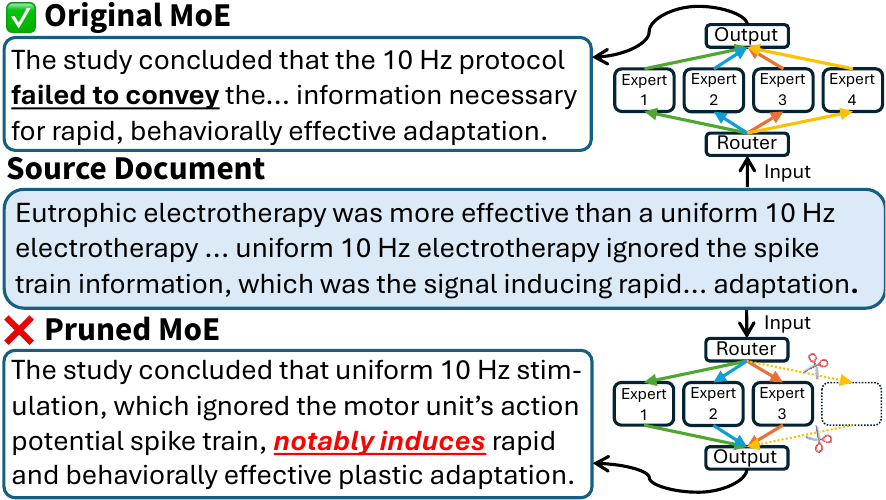}
    \caption{An example of hallucination in biomedical text summarization, where the text highlighted in red represents a term that inverts the original meaning of the source while it is still plausible.}
    \label{fig:motivation}
\end{figure}

A popular approach to mitigating this constraint is expert pruning, which eliminates redundant experts based on an informativeness (i.e., saliency) criterion estimated from a small calibration dataset~\citep[\textit{inter alia}]{chen2022taskspecificexpertpruningsparse,muzio2024seermoesparseexpertefficiency,lu-etal-2024-experts}.
Among these approaches, domain-specific expert pruning has been proposed to tailor the compressed model for a particular field~\citep{dong2025domainspecific}, such as biomedicine.
However, previous studies primarily evaluate pruning methods on downstream benchmark performance, or \textbf{utility}.
Consequently, they overlook the \textbf{reliability} of the resulting compressed models. This includes factual consistency, faithfulness, and hallucination prevention~\citep{10.1145/3571730,pal-etal-2023-med}.
This oversight poses serious risks particularly in high-stakes domains such as biomedicine, where factual errors can lead to critical real-world failures~\citep{10.1145/3531146.3533088,Moor2023}.

Furthermore, benchmark utility does not always align with factual reliability. \citet{chrysostomou-etal-2024-investigating} indicate that pruning dense models can result in \textbf{reduced} hallucinations. However, unlike dense model pruning, expert pruning completely removes entire weight matrices, fundamentally altering the model architecture.
This forces the model to rely on fallback experts; these alternatives may generate contextually fluent text while silently injecting subtle factual inaccuracies (see Figure \ref{fig:motivation}). 
Therefore, the relationship between utility and reliability in MoE pruning remains unclear.

In this paper, we investigate how domain-specific expert pruning affects \textit{both utility and reliability} in high-stakes settings.
Focusing on the biomedical domain, we evaluate four MoE models, six pruning methods, and multiple pruning ratios.
Our evaluation spans generation and classification tasks, comparing in-domain behavior with complementary general-domain analysis.
This framework examines whether pruning degrades performance, if utility and reliability remain coupled, where degradation begins to appear, and how these outcomes vary across task types and domains.
Our contributions are as follows:
\begin{itemize}
    \item We provide the first systematic study of how expert pruning affects the factual reliability of MoE models in a high-stakes setting, with a primary focus on the biomedical domain.

    \item We characterize how utility and reliability change across pruning ratios, model families, and pruning strategies, showing that moderate in-domain pruning remains robust while degradation becomes more pronounced at extreme pruning ratios and under domain shift.

    \item We show that the safety of MoE compression is strongly task- and domain-dependent, and that evaluating pruned MoE models using utility alone is insufficient for high-stakes deployment without explicit reliability assessment.
\end{itemize}

\section{Related Work}
\label{sec:related_work}

\paragraph{Reducing Memory Footprint in MoE Models.}
Efforts to reduce the memory footprint of MoE models span several approaches, including weight quantization~\citep[\textit{inter alia.}]{li2024quantmoebenchexaminingposttrainingquantization,huang2025mixture,pmlr-v267-chen25aa}, expert merging~\citep[\textit{inter alia.}]{he-etal-2023-merging,zhang-etal-2025-diversifying,zhou-etal-2025-dropping}, and expert pruning~\citep{kim2021scalableefficientmoetraining,lu-etal-2024-experts,chen-etal-2025-eac,NEURIPS2025_511c7fd6}.

Quantization techniques minimize network storage requirements by reducing parameter bit-width.
Examples include methods that vary bit-widths based on structural sensitivity \citep{li2024quantmoebenchexaminingposttrainingquantization,huang2025mixture}, optimize calibration \citep{pmlr-v267-chen25aa}, or employ extreme 1-bit compression \citep{yuan23c_interspeech,MLSYS2024_c74b6248}.
Quantization reduces numerical precision rather than altering network topology, making it orthogonal to expert merging and pruning~\citep{lu-etal-2024-experts}.

Expert merging mitigates memory constraints by mathematically blending parameter matrices into a unified matrix~\citep{he-etal-2023-merging,li2024merge,chen2025retrainingfree}, while expert pruning permanently removes a subset of redundant or low-saliency experts based on calibration data~\citep{muzio2024seermoesparseexpertefficiency,hu-etal-aaai26,lasby2026reap}.
Expert merging introduces irreducible errors by eliminating the ability of the router to maintain fine-grained, independent control over experts~\citep{dong2025domainspecific,lasby2026reap,liu2026evoesapnonuniformexpertpruning}.
Conversely, expert pruning preserves the original functional topology and routing independence, demonstrating superior performance at high compression rates~\citep{NEURIPS2025_511c7fd6,lasby2026reap}. Therefore, we focus exclusively on expert pruning.

\paragraph{Expert Pruning.}
Expert pruning methods are generally categorized into two paradigms: those that require subsequent fine-tuning and those that operate entirely without training (i.e., post-training or one-shot pruning)~\citep{lu-etal-2024-experts}. 
Methods requiring training remove experts based on routing statistics or regularization, subsequently applying gradient-based optimization to recover the resulting performance degradation~\citep{kim2021scalableefficientmoetraining,chen2022taskspecificexpertpruningsparse,muzio2024seermoesparseexpertefficiency,yang-etal-2024-moe}. 

In contrast, training-free expert pruning removes redundant experts using one-shot calibration data, bypassing parameter updates.
Prominent examples estimate expert saliency through reconstruction loss~\citep{lu-etal-2024-experts}, domain-specific demonstrations~\citep{dong2025domainspecific}, expert-selection frequency~\citep{chen-etal-2025-eac}, or advanced functional criteria to preserve routing independence without gradient updates~\citep{hu-etal-aaai26,lasby2026reap,liu2026evoesapnonuniformexpertpruning}.
We focus on this training-free paradigm because it provides a lightweight solution that facilitates the efficient on-device deployment of massive MoE architectures.
This approach also enables isolated examination of pruning effects without confounding variables from fine-tuning.

\paragraph{Model Compression and Factual Reliability.}
The relationship between model compression and the factual reliability of LLMs has been extensively investigated across dense model pruning, model merging, and weight quantization.
\citet{chrysostomou-etal-2024-investigating} find that dense model pruning reduces hallucinations in abstractive summarization by forcing reliance on source documents.
Other studies show that it degrades overall trustworthiness~\citep{pmlr-v235-hong24a} and disrupts internal activation features necessary for robust lie detection~\citep{fu-etal-2025-pruning}.
Therefore, dense model pruning yields task-dependent outcomes.
While studies on merged models remain scarce, \citet{NEURIPS2025_a3948805} demonstrate its utility as a parameter-level conflict-resolution strategy to harmonize helpfulness, honesty, and harmlessness.
For weight quantization, previous work~\citep{pmlr-v235-hong24a,singh-sajjad-2025-interpreting} demonstrates that moderate bit-width reduction generally preserves trustworthiness and internal calibration.
However, quantized models become susceptible to deceptive prompts, even while retaining truthful internal representations~\citep{fu-etal-2025-quantized}, and exhibit degraded faithfulness when generating natural language self-explanations~\citep{wang2026largelanguagemodelsexplain}.
Consequently, weight quantization presents a delicate trade-off between numerical efficiency and semantic precision.

Despite extensive research across these methodologies, \textit{the impact of expert pruning on factual reliability remains unexplored.}
This work provides the first systematic exploration into how expert pruning affects the reliability of MoE models, offering critical insights for their safe deployment.

\section{Training-free Domain-specific Expert Pruning} \label{sec:expert_pruning}

\subsection{MoE Pruning Framework}
Consider an MoE model with $L$ layers, where each layer $l \in \{1, \dots, L\}$ contains a set of $N$ experts, $\{E_{1}^{l}, \dots, E_{N}^{l}\}$.
For a given pruning ratio $p$, the objective is to retain a subset of $M$ experts per layer, where $M=(1-p)N$, and discard the remaining $N-M$ experts.
In this work, we focus on domain-specific expert pruning~\citep{dong2025domainspecific}, compressing an MoE model to preserve utility on a domain of interest (here, the biomedical domain).

To achieve this, the pruning process utilizes a calibration set $\mathcal{C}$ sampled from the domain of interest.
For each sample in $\mathcal{C}$ with a sequence length of $T$ tokens, a saliency metric (defined in \S\ref{subsec:saliency_metrics}) evaluates expert importance. After calculating the average importance across $\mathcal{C}$, experts with the lowest scores are removed. During inference, the router is restricted to selecting from the remaining $M$ experts.

\subsection{Saliency Metrics}
\label{subsec:saliency_metrics}
To quantify the importance of each expert, we consider six distinct metrics ranging from random baselines to state-of-the-art methods. Our aim is to investigate how different importance metrics affect domain utility and factual reliability.
Specifically, we first use a stochastic baseline (Random) and then advance through a progression of data-driven complexity: from static measures (Frequency), to dynamic activation-based approaches, and finally to recent, context-aware formulations.

\paragraph{Random.}
The random pruning metric samples $M$ experts for retention from a uniform distribution. This approach serves as a weak baseline excluding importance estimation and data-driven signals.

\paragraph{Frequency.}
This metric measures how often the router selects an expert~\citep{chen2022taskspecificexpertpruningsparse}. It assigns equal weight to every activation event, regardless of the magnitude assigned by the routing algorithm. For an expert $E_{i}^{l}$, the score is the count of activations: $\sum_{t=1}^{T} \mathbb{I}(g_{i,t}^{l}>0)$, where $\mathbb{I}(\cdot)$ is the indicator function that returns one for positive gating values and zero otherwise.

\paragraph{Gate.}
This metric assesses expert importance based on routing activation magnitude~\citep{chen2022taskspecificexpertpruningsparse}. Unlike Frequency, which treats all activations equally, the gating metric identifies experts prioritized by the router. For an expert $E_{i}^{l}$ in layer $l$, the score is the sum of gating values across all tokens: $\sum_{t=1}^{T} g_{i,t}^{l}$, where $g_{i,t}^{l}$ represents the gating value of the $i$-th expert for the $t$-th token at layer $l$.

\paragraph{EAN.}
Expert Activation Norm~\citep{jaiswal2025findingfantasticexpertsmoes} assesses importance by accumulating the norms of intermediate activations produced by an expert. The saliency score for expert $E_{i}^{l}$ is the $L_{2}$ norms of outputs across active tokens: $\sum_{t=1}^{T} \mathbb{I}(g_{i,t}^{l}>0)\cdot\|E_{i,t}^{l}(h_{t}^{l})\|_{2}$. Here, $E_{i,t}^{l}(h_{t}^{l})$ denotes the output vector of the $i$-th expert for the hidden state $h_{t}^{l}$ at layer $l$. This method favors experts that produce high-magnitude transformations.

\paragraph{EASY-EP.}
Expert Assessment with Simple Yet-effective scoring for Expert Pruning~\citep{dong2025domainspecific} optimizes importance estimation by coupling the output magnitude of an expert with the token-level contribution to representation shift. This metric posits an expert is essential only when the output magnitude is large and the transformation induces a considerable reorientation of the hidden state. The saliency score is the product of output-aware importance $c_{i,t}^{l}$ and expert-level token contribution $s_{t}^{l}$, aggregated as $\sum_{t=1}^{T} c_{i,t}^{l}\cdot s_{t}^{l}$. The output-aware importance $c_{i,t}^{l}$ multiplies the routing gate value by the $L_{2}$ norm of the expert output, formulated as $c_{i,t}^{l}=g_{i,t}^{l}\|E_{i,t}^{l}(h_{t}^{l})\|_{2}$. The token contribution $s_{t}^{l}$ is defined as $1-\text{Sim}(h_{t}^{l},\tilde{h}_{t}^{l})$, where $\text{Sim}$ denotes cosine similarity between hidden representations preceding ($h_{t}^{l}$) and following ($\tilde{h}_{t}^{l}$) the expert module.\looseness=-1

\paragraph{REAP.}
Router-weighted Expert Activation Pruning \citep{lasby2026reap} extends EAN by integrating gating values and normalizing by activation frequency.
This metric isolates experts with the most substantial per-token contribution via the formulation: $\frac{\sum_{t=1}^{T} g_{i,t}^{l} \|E_{i,t}^{l}(h_{t}^{l})\|_{2}}{\sum_{t=1}^{T} \mathbb{I}(g_{i,t}^{l} > 0)}.$
This approach ensures that the importance score reflects the mean contribution per activation, identifying experts that maintain influence regardless of selection frequency.

\section{Experimental Setup}

To investigate the empirical relationship between expert pruning and factual consistency, the evaluation framework comprises two distinct dimensions: \textbf{utility}, reflecting downstream task performance, and \textbf{reliability}, quantifying hallucination frequency.
This joint analysis reveals whether configurations that maintain in-domain utility concurrently preserve reliability, or if reliability degrades more rapidly than utility as the pruning ratio increases.\footnote{Please see Appendix \ref{appendix:impl_details} for implementation details.}\looseness=-1

\subsection{Models}
We use four instruction-tuned MoE models: GPT-OSS 20B~\citep[GPT-OSS, 32 experts]{openai2025gptoss120bgptoss20bmodel}, Qwen3 30B Instruct 2507~\citep[Qwen3, 128 experts]{yang2025qwen3technicalreport}, Nemotron 3 Nano 30B \citep[Nemotron3, 128 experts]{nvidia2025nvidianemotron3efficient}, and Qwen3.6 35B \citep[Qwen3.6, 256 experts]{qwen36_35b_a3b}.
This selection offers varying expert granularity and architectures such as the hybrid Mamba-Transformer Nemotron3.

\subsection{MoE Pruning}
We utilize MedINST~\citep{han-etal-2024-medinst} as the calibration set $\mathcal{C}$ for expert selection.
MedINST is a comprehensive meta-dataset of biomedical instructions with diverse tasks, providing an ideal foundation to capture representative domain-specific activations.
The calibration set contains 128 randomly sampled demonstrations from the training subset of MedINST, following the established practice in the pruning literature~\citep{williams-etal-2025-self}.
To mitigate sampling bias, we run three independent calibration experiments for each configuration.
Furthermore, to systematically evaluate the impact of structural compression, we assess each pruning strategy across pruning ratios in 12.5\% increments.

\subsection{Utility}
\label{subsec:utility}
\paragraph{Generation Tasks.}
We employ a MedINST evaluation subset requiring full-text generation.\footnote{This subset does not overlap with the calibration set, ensuring a fair comparison.}
The selected categories are summarization (SUM), machine translation (MT), question answering (QA), named entity recognition (NER), named entity disambiguation (NED), relation extraction (RE), coreference resolution (COREF), and event extraction (EE).
We compute zero-shot ROUGE-L~\citep{lin-2004-rouge} for SUM, chrF++~\citep{popovic-2017-chrf} for MT, and F1 for all other tasks.

\paragraph{Classification Tasks.}
To complement the generative evaluation, we follow \citet{williams-etal-2026-compressing} to assess discriminative utility using the MultiMedQA benchmark~\citep{Singhal2023}.
The benchmark comprises several multiple-choice QA tasks: PubMedQA~\citep{jin-etal-2019-pubmedqa}, MedQA~\citep{app11146421}, and relevant subsets from MMLU (anatomy, clinical knowledge, college medicine, medical genetics, professional medicine, and college biology)~\citep{hendrycks2021measuring}. We measure zero-shot accuracy for all tasks.

\subsection{Reliability}
\label{subsec:reliability}
\paragraph{Generation Tasks.}
We measure semantic consistency using the Multi-XScience~\citep{lu-etal-2020-multi-xscience} and RCT~\citep{AMIA-summarization-2021} benchmarks.
Multi-XScience, a part of MedINST, involves generating a related work section based on an abstract and reference articles.
The RCT dataset contains randomized controlled trial reports and serves as a high-stakes test for medical summary accuracy.

We use an LLM-as-a-Judge framework~\citep{zheng2023judging} comprising three frontier models (\href{https://developers.openai.com/api/docs/models/gpt-5.4-mini}{\texttt{gpt-5.4-mini}}, \href{https://www.anthropic.com/claude-haiku-4-5-system-card}{\texttt{Claude Haiku 4.5}}, and \href{https://storage.googleapis.com/deepmind-media/Model-Cards/Gemini-3-1-Flash-Lite-Model-Card.pdf}{\texttt{Gemini 3.1 Flash-Lite}}) to minimize inter-model variability.
The evaluation includes two methods: \textbf{absolute judgment}, which classifies outputs as either faithful or hallucinated, and \textbf{relative judgment}~\citep{lango-dusek-2023-critic,chrysostomou-etal-2024-investigating}, which compares summaries from source and pruned models across four criteria:
\begin{enumerate}
\item \textbf{Hallucinations} ($\downarrow$): Frequency of unsupported content by the source document.
\item \textbf{Omission} ($\downarrow$): Exclusion of critical source information.
\item \textbf{Repetition} ($\downarrow$): Presence of redundant text.
\item \textbf{Alignment} ($\uparrow$): Degree of semantic correspondence with the source document.
\end{enumerate}
For relative judgment, we report the preference rate, calculated as the proportion of instances favoring the pruned model over the source model, where a rate above 0.5 denotes preference for the pruned model. 
We also compute standard metrics such as ROUGE-L and BERTScore~\citep{Zhang2020BERTScore} to contrast lexical and semantic overlap with these judgments, following~\citet{chrysostomou-etal-2024-investigating}.

\paragraph{Classification Tasks.}
For discriminative reliability, we use three reasoning tasks from the Medical Domain Hallucination Test~\citep[MedHALT]{pal-etal-2023-med}: the False Confidence Test (FCT), the Fake Questions Test (Fake), and the None of the Above Test (NOTA).
The FCT provides a randomly suggested answer alongside the question to assess whether the model exhibits unwarranted certainty.
The Fake task presents nonsensical medical questions to evaluate whether the model can recognize invalid queries.
Finally, the NOTA task replaces the correct option with ``None of the above'' to test whether the model can reject incorrect information.
Collectively, these tests determine whether the model resists hallucination when a false answer is proposed, a query is fundamentally flawed, or a correct solution is absent.

\subsection{General-domain Analysis}
To complement our domain-specific evaluation, we monitor the impact of expert pruning on general-domain benchmarks.
This analysis determines whether the relationship between utility and reliability in the biomedical domain persists across a broader context.

\paragraph{Utility.}
We use four benchmarks: IFEval~\citep{zhou2023instructionfollowingevaluationlargelanguage} zero-shot accuracy for instruction-following, GSM8K~\citep{cobbe2021trainingverifierssolvemath} five-shot exact match for math reasoning, HumanEval~\citep{chen2021evaluatinglargelanguagemodels} zero-shot pass@1 for coding, and MMLU three-shot accuracy for general reasoning.

\paragraph{Reliability.}
We use Multi-News+~\citep{choi-etal-2024-multi-news} to measure the ability of the model to summarize multiple documents for a given topic.
We apply the same LLM-as-a-Judge protocol from  \S\ref{subsec:reliability}.

\section{Results}
\subsection{Biomedical Domain Utility}
\label{subsec:utility_results}
\begin{figure}[t]
    \centering
    \includegraphics[width=\linewidth, trim=0bp 2bp 0bp 0bp, clip]{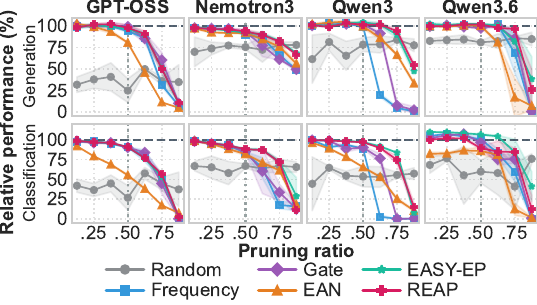}
    \caption{Downstream performance comparison across pruning ratios. Lines denote mean performance relative to the source baseline (100\%), with shaded areas indicating standard deviation across three random seeds. Task-specific results at a 50\% pruning ratio are available in Tables \ref{tab:utility_gen_target} and \ref{tab:utility_cls_target} in the Appendix.}
    \label{fig:ratio_comparison}
\end{figure}

Figure \ref{fig:ratio_comparison} shows the relative performance retention across various compression levels.

\begin{figure*}[t]
    \centering
    \includegraphics[width=0.95\linewidth, trim=0bp 5bp 0bp 0bp, clip]{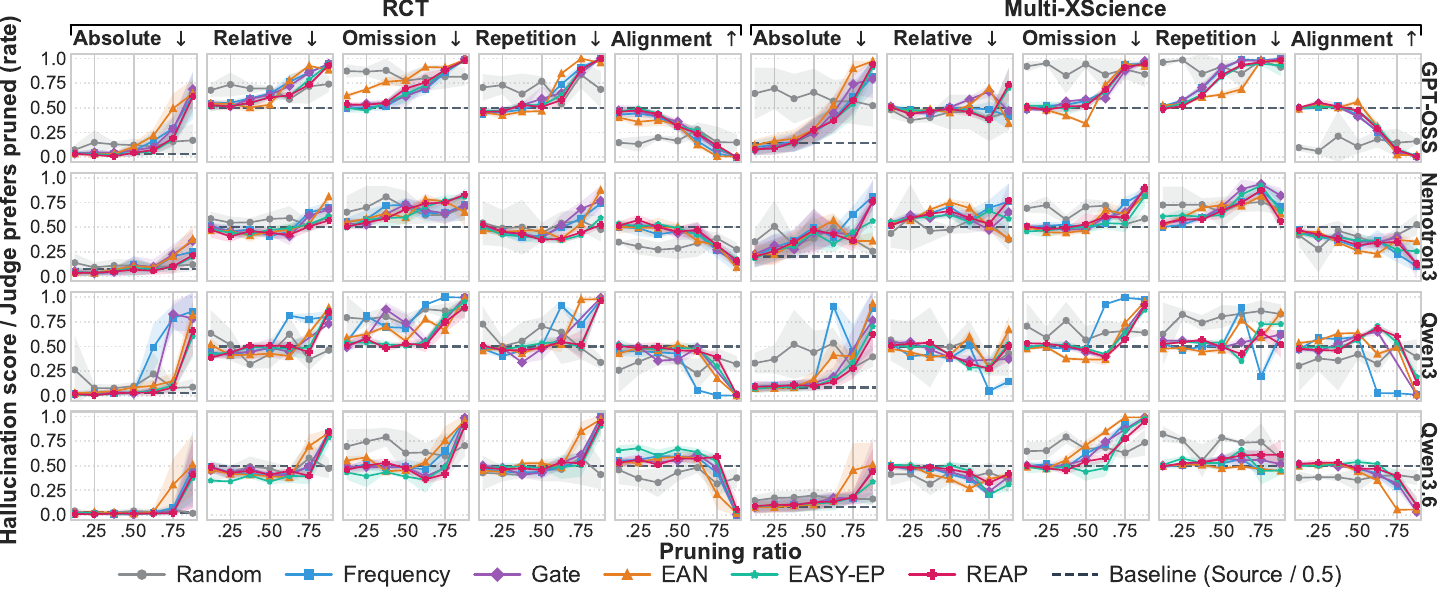}
    \caption{Biomedical reliability results across pruning ratios. Dashed lines denote baseline unpruned models for absolute scores, or the 0.5 preference threshold for relative comparisons.
    The average inter-annotator agreements (Fleiss' $\kappa$) are 0.52 for absolute judgments and 0.46 for relative judgments (moderate agreement).
    Tables~\ref{tab:agreement_absolute} and~\ref{tab:agreement_relative} in the Appendix provide agreement breakdowns. Figure \ref{fig:sum_ratio_results} in the Appendix shows the corresponding summarization metrics.\looseness=-1}
    \label{fig:reliability_bio}
\end{figure*}

\paragraph{Pruning Method.}
We observe that expert pruning strategies influence in-domain performance retention.
Random pruning causes severe performance degradation across all settings.
At a 50\% pruning ratio, a standard evaluation threshold in prior MoE pruning work~\citep{dong2025domainspecific,lasby2026reap}, it leads to relative utility drops ranging from 18.8\% for Qwen3.6 to 75.1\% for GPT-OSS on generation tasks.

In contrast, non-random strategies retain performance comparable to unpruned baselines.
Most of these strategies yield similar results, with a maximum variance of 5.2\% in generation tasks and 17.2\% in classification tasks.
However, EAN often lags behind other methods in classification tasks, underperforming the best data-driven metric (EASY-EP) by an average of 23.7\% across models, with the gap reaching 36.0\% in GPT-OSS.

This observation aligns with recent findings~\citep{dong2025domainspecific,lasby2026reap}.
Relying solely on raw activation norms forces EAN to overemphasize output scale instead of true token-level utility.
Because EAN does not consider the gating confidence of the router ($g_{i,t}^{l}$ in REAP) or the representational shift ($s_{t}^{l}$ in EASY-EP), it retains experts that produce large but contextually unhelpful transformations. This generates suboptimal utility.
Consequently, EAN experiences an earlier performance drop compared to other approaches at a 62.5\% pruning ratio.
Importantly, this earlier utility decline corresponds to an earlier onset of hallucination risks across models, as discussed in \S\ref{subsec:reliability_results}.\looseness=-1

\paragraph{Model Family.}
Pruning behavior also varies across expert granularities and model architectures.
Models with large expert pools, like Qwen3 and Qwen3.6, tolerate a 50\% pruning ratio well.
For Qwen3.6, all data-driven approaches achieve nearly 100\% performance retention on generation tasks, and EASY-EP even reaches 106.8\% on classification tasks.
In contrast, models with fewer experts, like GPT-OSS, suffer greater degradation because each expert holds more network capacity.
This vulnerability is especially evident on classification tasks, where no configuration exceeds 91.1\%.

Beyond expert count, architectural complexity introduces another vulnerability.
Nemotron3 models show a consistent performance decline, underperforming the baseline by at least 6.3\% (REAP).
We hypothesize that expert pruning disrupts the balance within this hybrid architecture, which combines Transformer attention, Mamba state-space layers, and MoE feed-forward networks.
This instability is critical in classification tasks, which rely on precise internal state propagation to form accurate decision boundaries.

\paragraph{Pruning Ratio.}
For generation tasks, performance remains robust up to a 50\% pruning ratio. 
Context-aware methods like EASY-EP and REAP often match or exceed the unpruned baseline at ratios up to 75\%.
For instance, at a 75\% pruning ratio, Qwen3.6 under REAP retains 96.7\% of its baseline utility (a negligible 3.3\% drop).
In contrast, classification tasks show a gradual performance decline before dropping severely.

\begin{figure}[t]
    \centering
    \includegraphics[width=\linewidth, trim=0bp 5bp 0bp 0bp, clip]{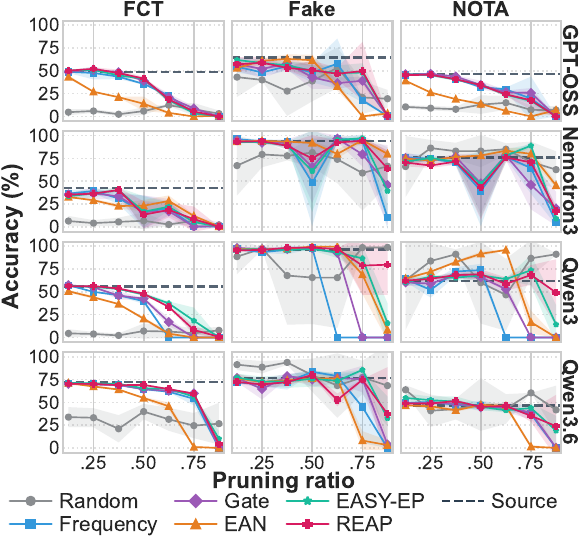}
    \caption{MedHALT evaluation results across pruning ratios and approaches. Dashed lines represent unpruned baseline model performances.}
    \label{fig:medhalt_bio}
\end{figure}

\subsection{Biomedical Domain Reliability}
\label{subsec:reliability_results}

\paragraph{Pruning Method.}
Most pruning methods preserve baseline reliability metrics up to moderate compression levels (a 50\% pruning ratio) in-domain.
However, we observe that Random and EAN pruning degrade earlier; Random pruning causes immediate degradation even under moderate compression.
On FCT classification, accuracy approaches near zero across all pruning ratios (Figure~\ref{fig:medhalt_bio}).
On generation tasks (Figure~\ref{fig:reliability_bio}), Random pruning causes severe omissions that standard summarization metrics often mask.
For instance, on RCT at a 37.5\% pruning ratio, Qwen3.6 under Random pruning maintains a ROUGE-L of .135 and BERTScore of .853 (vs. .134 and .853 for the baseline).
Yet, its relative alignment drops to .329 due to high omissions (.790).
EAN also exhibits early reliability degradation under moderate pruning compared to other data-driven approaches.
For example, on FCT, Qwen3 accuracy drops to 20.7\% under EAN at a 50\% ratio, compared to 47.2\% under EASY-EP.
\begin{observationbox}{Coherence vs. Completeness}
Random pruning can result in high omission and repetition rates even under moderate compression. Standard summarization metrics fail to identify cases where summaries remain structurally fluent but omit key information, creating a false impression of stability.
\end{observationbox}

\paragraph{Model Family.}
Echoing the utility findings (\S\ref{subsec:utility_results}), models with large expert pools, such as Qwen3 and Qwen3.6, exhibit remarkable stability up to moderate pruning ratios (50\%).
This robustness particularly holds under context-aware pruning methods (EASY-EP, REAP), where both generative and discriminative benchmarks remain stable and match baseline performance.

In contrast, models with fewer experts show earlier degradation.
GPT-OSS exhibits reliability degradation at a 50\% pruning ratio.
Under EASY-EP at this ratio, GPT-OSS suffers from substantial factual consistency decay on both RCT and Multi-XScience, despite minimal decline in both task-specific and overall utility metrics.
For instance, on Multi-XScience, the model suffers from high repetitions (.815), alongside an increase in the absolute hallucination rate to 17.3\% (compared to 10.5\% at the 37.5\% ratio).
However, task-specific summarization metrics remain stable. ROUGE-L and BERTScore are .102 and .817 on Multi-XScience (vs. .100 and .816 for the baseline).
Likewise, overall generation utility retains 96.9\% of baseline performance (a minor 3.1\% drop), demonstrating that \textit{factual reliability decay can occur even with minimal impact on standard task-specific and overall utility measurements.}
\begin{observationbox}{MoE Capacity Bottleneck}
Under moderate pruning ratios ($\approx 50\%$), models with smaller expert pools can exhibit factual reliability degradation earlier than models with larger expert pools.
This decline occurs before standard utility metrics begin to degrade.
\end{observationbox}

Nemotron3, while stable under low pruning ratios of 25\% or lower, undergoes gradual reliability degradation at the moderate stage of 50\%.
On RCT under EASY-EP, it exhibits increased information omissions (.598).
On Multi-XScience, it experiences a decline in relative alignment (.364) accompanied by elevated relative hallucinations (.625) and repetitions (.624).
Yet, summarization metrics on Multi-XScience remain largely unaffected: ROUGE-L is .137 and BERTScore is .838, compared to the baseline scores of .148 and .845.
This stability hides the factual consistency decay as the absolute hallucination rate rises to 44.7\% from 20.1\% for the source model.
However, overall generation utility successfully detects this degradation, underperforming the baseline by 7.9\% at 50\% pruning.
This suggests that physical model constraints (e.g., the hybrid MoE-Mamba-Attention topology of Nemotron3) drive a coupled, parallel decay in overall utility and reliability, even when task-specific metrics remain stable.

\paragraph{Pruning Ratio.}
Extreme pruning (62.5\% or higher) increases reliability risks across all models, mirroring utility drops.
However, the threshold for reliability preservation depends on the task.
For instance, tasks like Fake, NOTA, and RCT often maintain baseline reliability even at 75\% compression.
Conversely, tasks like FCT and Multi-XScience exhibit earlier, gradual degradation.
We attribute this divergence to different reasoning demands.
Simple classification (Fake and NOTA) and single-source extraction (RCT) may rely on broad semantic matching shared across many experts.
In contrast, precise fact retrieval (FCT) and multi-source synthesis (Multi-XScience) depend on specific experts, making them vulnerable to pruning.

Crucially, extreme compression decouples utility and reliability on Multi-XScience for GPT-OSS, Qwen3, and Qwen3.6.
For instance, under EASY-EP at 75\% pruning, the relative hallucination rate for Qwen3.6 decreases to .204 from .507 at 50\% pruning.
Yet, the corresponding absolute rate increases from 11.9\% to 17.9\%.
This apparent relative improvement is due to an artifact of summary collapse.
A summary length audit reveals that at 75\% pruning, average lengths drop from baselines for Qwen3.6 (204.9 to 97.4 words), Qwen3 (367.8 to 262.7), and GPT-OSS (382.9 to 279.8).
Consequently, relative omission rates surge above .630 for these models.
The shorter the summary, the lower the absolute number of errors it will contain, lowering the probability of incurring a penalty during relative hallucination evaluation.
However, the absolute hallucination rate rises because the actual summary consists largely of ungrounded fabrications.
Standard summarization metrics mask this collapse; ROUGE-L and BERTScore remain largely stable (e.g., Qwen3.6 records .122 and .833, compared to .152 and .846 at 50\% pruning).
\begin{observationbox}{Summary Collapse}
Extreme pruning artificially lowers relative hallucination rates while standard summarization metrics remain stable.
This stems from summary collapse; degenerate outputs omit critical information and make fewer assertions, rendering them less faithful to the source material.
\end{observationbox}

\subsection{General-domain Analysis}
\label{subsec:general_results}
\begin{figure}
    \centering
    \includegraphics[width=0.95\linewidth, trim=0bp 2.1bp 0bp 0bp, clip]{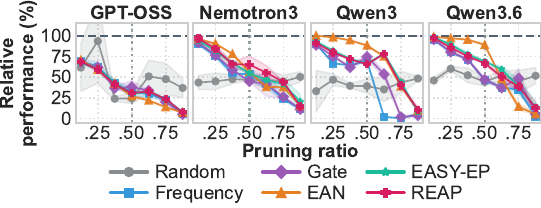}
    \caption{General-domain utility across pruning ratios. Lines and shaded areas denote the mean performance relative to the unpruned source baseline (100\%) and standard deviation, respectively. Table \ref{tab:utility_general} in the Appendix provides task-specific results at 50\% pruning.}
    \label{fig:ratio_comparison_general}
\end{figure}

\paragraph{Utility.}
Figure \ref{fig:ratio_comparison_general} presents the results on general-domain benchmarks.
Across model families, expert pruning causes an immediate performance decline in the general domain, with overall utility dropping steadily as the pruning ratio increases.
This gradual decay contrasts with in-domain utility, where performance remains highly robust up to 50\%--75\% pruning before collapsing suddenly.
GPT-OSS is the least robust and shows an immediate average decline of 32.4\% even at 12.5\% pruning probably due to its small expert pool.

When comparing pruning methods, EAN occasionally beats others up to a 50\% ratio, except for GPT-OSS.
Because EAN selects experts by magnitude without domain-specific weighting, it preserves general capabilities at the expense of in-domain alignment.
This trade-off becomes evident when examining expert selection overlap.
At 50\% pruning, the overlap between EAN and other data-driven methods is low, peaking at .653 for Qwen3.6 (with the exception of Nemotron3 at .714).
In contrast, pairwise overlap among the remaining methods is higher with a minimum of .753.
Ultimately, this divergence occurs because domain-specific calibration inherently removes general-purpose experts that rarely activate during biomedical tasks.
Our multi-domain calibration analysis in \S\ref{sec:analysis} further corroborates this finding.

\begin{figure}
\centering
\includegraphics[width=0.98\linewidth, trim=0bp 5bp 0bp 0bp, clip]{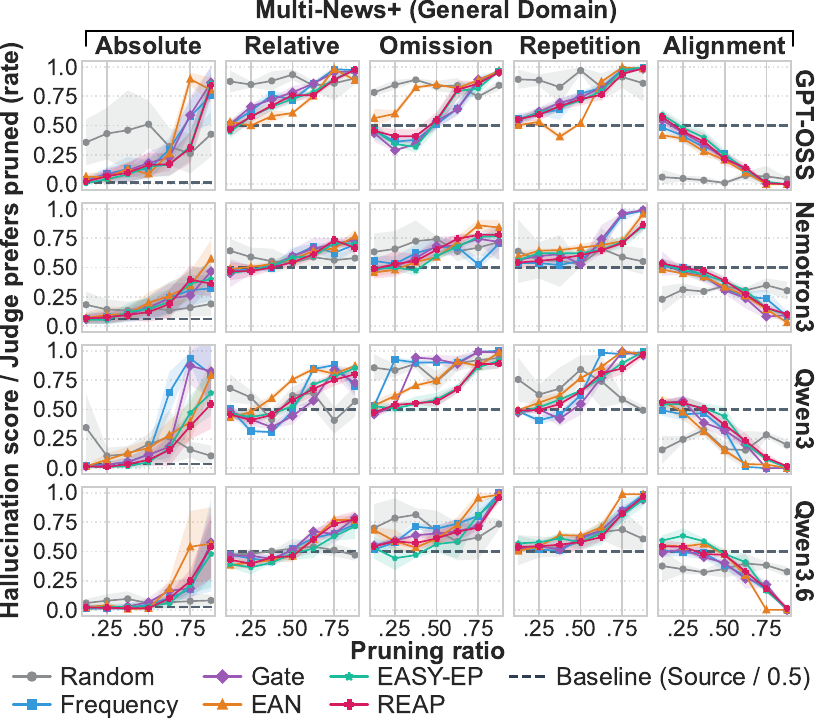}
\caption{General-domain reliability (Multi-News+) across pruning ratios.
Dashed lines denote unpruned baselines for absolute scores and the 0.5 preference threshold for relative comparisons.
The inter-annotator agreements (Fleiss’ $\kappa$) are 0.58 for absolute judgments and 0.44 for relative judgments (moderate agreement).
}
\label{fig:reliability_general}
\end{figure}

\paragraph{Reliability.}
Hallucination rates on Multi-News+ increase monotonically with pruning ratio (Figure \ref{fig:reliability_general}).
For non-random methods, Spearman correlation coefficients for absolute and relative hallucination rates span $\rho = 0.88$ to $0.93$ and $\rho = 0.73$ to $0.87$, respectively.
Crucially, this degradation occurs much earlier than in the in-domain setting.
Unlike in-domain reliability, which remains largely stable through 50\%--62.5\% pruning (and up to 75\% for Qwen3 and Qwen3.6) as observed in Figure \ref{fig:reliability_bio}, Multi-News+ hallucination rates climb steadily.
For example, GPT-OSS reaches an 11.1\% mean absolute hallucination rate at just 37.5\% pruning, diverging from a 1.3\% baseline.
Only Qwen3.6 remains relatively robust until 62.5\% pruning, where it reaches 12.2\% (baseline 2.7\%).

By contrast, summarization metrics remain stable.
Both ROUGE-L and BERTScore stay within 2\% of the unpruned baseline for all models up to 62.5\% pruning (Figure \ref{fig:sum_ratio_results} in the Appendix).
Because ROUGE-L measures the longest common subsequence and ignores repetition, and BERTScore rewards topically plausible yet unfaithful text, both metrics obscure the underlying degradation of reliability.
This decoupling constitutes another form of summary collapse driven by domain shift.
At 62.5\% pruning, models tend to produce repetitive text with increasing relative repetition rates (e.g., 77.1\% for Qwen3 under EASY-EP).
This triggers a collapse in lexical diversity, measured via Type-Token Ratio (TTR).
For instance, from an unpruned baseline of .765, the TTR of Qwen3 drops to .731 at 62.5\% pruning and falls sharply to .638 at 75\%.
Except for Qwen3.6, this repetition inflates mean summary length by up to 149 words without improving faithfulness.

\section{Analysis}
\label{sec:analysis}
\paragraph{Correlation between LLM-as-a-Judge and Human  Evaluation.}
To validate the reliability of our evaluation framework, we conduct a manual annotation study on 30 samples from RCT and Multi-XScience.
Three human evaluators annotate each sample for comparison against the judgments of our three LLM judges. We stratify the reliability agreement (Cohen's $\kappa$) by pruning ratio windows: Low ($p \le 25.0\%$), Moderate ($37.5\% \le p \le 50.0\%$), and Extreme ($p \ge 62.5\%$), as shown in Table~\ref{tab:agreement_by_window}.

\input{table/human_llm_agreement}

Under extreme pruning ratios, we find substantial agreement across all groups (e.g., human-LLM absolute agreement reaches 82.1\% with $\kappa = 0.644$). Severe quality degradation (e.g., obvious hallucinations, omissions, and repetitions) drives this straightforward evaluation for both humans and LLMs.
Under low and moderate pruning, however, the differences are subtle.
Relative preference exhibits substantial variance across both windows ($\kappa = 0.238$ and $\kappa = -0.012$ for human-LLM).
This variance reflects the inherent difficulty of choosing between outputs of similar quality before severe degradation occurs.
For absolute judgments, agreement is high during moderate compression ($\kappa = 0.738$ for human-LLM) but diminishes under low compression ($\kappa = 0.006$), suggesting that LLM judges diverge from humans when evaluating near-baseline outputs.
Overall, the strong alignment between human and LLM judges in detecting severe degradations justifies our automated evaluation setup for generation reliability assessment.\looseness=-1

\input{table/ablation}

\paragraph{Multi-domain Calibration.}

Given the substantial impact of calibration data domains on utility and reliability, we compare biomedical-only calibration against dual-domain calibration, which additionally uses general-domain calibration data.\footnote{We use 128 Dolci-Instruct-SFT-No-Tools~\citep{olmo2025olmo3} samples as general-domain calibration data.} Following \citet{dong2025domainspecific}, we compute a saliency metric (\S\ref{subsec:saliency_metrics}) for each domain and average them.
We analyze these calibration behavior at moderate (50\%) and extreme (75\%) pruning ratios using Qwen3.6 with EASY-EP as this combination represents the most effective pruning configuration.

As shown in Table~\ref{tab:ablation_results}, dual-domain calibration mitigates general utility degradation, boosting it from .387 to .597 at 75\% pruning.
Yet, this improvement incurs a trade-off in biomedical utility (e.g., at 75\% pruning, MultiMedQA drops from .665 to .563).
For reliability, dual-domain calibration performs comparably to the biomedical-only method at 50\% pruning.
It maintains low absolute hallucination rates, degrading by a maximum of .029 on Multi-News+ compared to the unpruned baseline, and yields similar pairwise preference win-rates.
Conversely, under extreme 75\% pruning, the biomedical-only calibration is more robust.
Under dual calibration, absolute hallucination rates surge across tasks, doubling the rates of MedINST (.358 vs.\ .179 on Multi-XScience).
Pairwise metrics confirm this degradation: the 75\% dual-pruned model exhibits higher relative hallucinations and repetitions.
This outcome suggests that while multi-domain calibration helps balance general and specialized utility under moderate pruning, specialized in-domain calibration is critical for preserving factual reliability during extreme compression.

\paragraph{Quantization.}

While quantization is orthogonal to expert pruning (\S\ref{sec:related_work}), we examine whether utility and reliability trends remain consistent in quantized models. Specifically, we apply 4-bit GPTQ~\citep{frantar2023optq} weight quantization to the Qwen3.6 EASY-EP 50\% pruned model.
As Table \ref{tab:ablation_results} indicates, we observe no substantial utility difference between the unquantized and quantized configurations: the quantized model retains a biomedical utility of .543 on generation and .829 on classification tasks (compared to .554 and .838 for the unquantized counterpart).
It also preserves .676 of general-domain utility, a minor decrease from the unquantized score of .689.

For reliability, quantization generally preserves consistency on biomedical benchmarks.
For instance, it maintains absolute hallucination rates of .017 on RCT and .125 on Multi-XScience, against unquantized rates of .015 and .121.
However, in the general domain, quantization increases absolute hallucinations (.078 vs.\ .031) and elevates relative errors. Hallucination, omission, and repetition rates rise to .547, .553, and .620 respectively, while relative alignment drops from .647 to .410.
These results demonstrate that while expert pruning and post-training weight quantization can combine to achieve additional compression without compounding in-domain performance loss, quantization can increase reliability risks in cross-domain settings.

\paragraph{Qualitative Analysis.}

\begin{table}[t]
\centering
\renewcommand{\arraystretch}{0.8}
\setlength{\aboverulesep}{1.3pt}
\setlength{\belowrulesep}{1.3pt}
\setlength{\tabcolsep}{3pt}
\resizebox{\linewidth}{!}{
\begin{tabular}{lllccc}
\toprule
& \textbf{Source Context} & \textbf{Pruned Output} & \textbf{Cl.} & \textbf{Gem.} & \textbf{GPT} \\
\midrule
\textbf{1.} & 3.8 ng/ml & 3.8 \textbf{mg/ml} & 1.0 & 0.0 & 0.0 \\
\textbf{2.} & 179 randomized & \textbf{70-9} randomized & 1.0 & 1.0 & 0.0 \\
\textbf{3.} & (abstract) & \texttt{Given ..., write...} & 1.0 & 1.0 & 0.0 \\
\bottomrule
\end{tabular}
}
\caption{Output error examples on RCT and LLM judge absolute evaluations (1.0: unfaithful, 0.0: faithful).}
\label{tab:qualitative_examples}
\end{table}

To investigate reliability degradation, we analyze RCT generated summaries (Table~\ref{tab:qualitative_examples}).
We identify three primary failure modes: (1) \textit{Unit hallucinations}, swapping measurement units (e.g., Qwen3 generating mg/ml instead of ng/ml at 25\% Gate); (2) \textit{Numerical errors}, introducing factual mistakes (e.g., GPT-OSS reporting 179 patients as 70-9 at 75\% Frequency); and (3) \textit{Instruction leakage}, repeating the task template under extreme pruning (e.g., Qwen3.6 at 87.5\% EAN).
Evaluating these outputs reveals the varying sensitivities of absolute LLM judges. While Claude identifies most errors, Gemini and GPT overlook the unit swap in Case 1.
This error notably occurs at a low 25\% pruning ratio while utility remains intact. This establishes that evaluating pruned models on utility alone is often inadequate for high-stakes deployment; direct reliability assessments are critical to detect factual failures before utility degrades.

\section{Conclusion}
This paper presents the first comprehensive study on the impact of expert pruning in MoE models on both benchmark utility and factual reliability within the high-stakes biomedical domain.
Our findings highlight that utility is not always indicative of reliability.
During domain shift or extreme pruning, models experience summary collapse and an increase in hallucination rates before standard utility metrics register a decline. In high-stakes domains, this trade-off between utility and reliability requires careful consideration through comprehensive in-domain and general evaluation. 
Finally, while multi-domain calibration and quantization offer benefits, they introduce reliability trade-offs under high compression or cross-domain settings.

\section*{Acknowledgment}
We would like to thank Mingzi Cao, Vynska Amalia Permadi, and Samuel Lewis-Lim for their annotation support.
We also appreciate initial guidance on hallucination evaluation from Timothee Mickus.
AY is supported by the Engineering and Physical Sciences Research Council (EPSRC) [grant number EP/W524360/1] and the Japan Student Services Organization (JASSO) Student Exchange Support Program (Graduate Scholarship for Degree Seeking Students).

\bibliography{custom,anthology-1,anthology-2}
\bibliographystyle{acl_natbib}

\clearpage
\appendix
\section{Implementation Details}
\label{appendix:impl_details}
\paragraph{Software.}
We utilize Hugging Face (HF) datasets~\cite[v3.6.0]{lhoest-etal-2021-datasets} for data preprocessing, alongside HF transformers~\cite[v5.5.4]{wolf-etal-2020-transformers}, FlashAttention-2~\cite[v2.7.4]{dao2024flashattention}, and PyTorch~\cite[v2.10.0]{10.1145/3620665.3640366} for the pruning framework.
We employ lm-evaluation-harness~\cite[v0.4.10]{eval-harness} for IFEval, GSM8K, HumanEval, and MMLU evaluation.
For MedHALT, we use the official repositories~\citep[Commit: bd4408a]{pal-etal-2023-med}.\footnote{\url{https://github.com/medhalt/medhalt/tree/bd4408a16aff36626e934aa5b012edd9fa7b6194}}
For HaluEval, we adopt the dataset from the official HF repository.\footnote{\url{https://huggingface.co/datasets/pminervini/HaluEval}}
For RCT, we obtain the dataset from the official repository~\citep[Commit: de10c27]{AMIA-summarization-2021}.\footnote{\url{https://github.com/bwallace/RCT-summarization-data/tree/de10c2712873efa1733859f2d7113af60427d7b2}}
Similarly, for Multi-News+, we load the dataset from the official repository~\citep[Commit: e347c8e]{choi-etal-2024-multi-news}.\footnote{\url{https://github.com/c-juhwan/multi_news_plus/tree/e347c8eedb78a09b4971bc0011a8865e11dfafdb}}
Due to API cost constraints during LLM-as-a-Judge evaluation, we randomly subsample 50 documents from the Multi-News+ test set for our evaluation.
For MedINST, we execute the official script~\citep[Commit: dc13b2b]{han-etal-2024-medinst}.\footnote{\url{https://github.com/aialt/MedINST/blob/dc13b2be29cdaae01cddceb6e4134e7dc6df1134/evaluation.py}}
For absolute judgment, we leverage the pre-configured \texttt{HallucinationMetric} of DeepEval~\citep[v3.9.6]{Ip_deepeval_2026}.
Inference is executed using the vLLM~\citep[v0.19.1]{10.1145/3600006.3613165} engine for faster inference.
For post-training weight quantization analysis, we apply 4-bit GPTQ using the \texttt{llmcompressor} library~\citep[v0.12.0]{Red_Hat_AI_and_vLLM_Project_LLM_Compressor_2024}.

\paragraph{Hardware.}
We use a single NVIDIA A100 80GB GPU with CUDA 12.9 for experiments.
Notably, the expert pruning process completes within 775 seconds (using the largest model: Qwen3.6 with the context-aware EASY-EP method) and introduces no substantial computational overhead.

\paragraph{Prompt Templates.}
We prioritize official prompt templates to maintain standardized evaluation conditions, while using custom templates only as required.
We use the default prompt templates for MultiMedQA, IFEval, GSM8K, HumanEval, and MMLU provided by lm-evaluation-harness.
For MedINST, MedHALT, and HaluEval, we utilize their official prompt templates.
For summarization tasks, including RCT and Multi-News+, we apply the custom prompt templates listed in Appendix \ref{appendix:prompt}.
For LLM-as-a-Judge and human evaluation, we employ the off-the-shelf DeepEval template for absolute judgment and a custom prompt based on \citet{chrysostomou-etal-2024-investigating} for relative judgment; both are documented in Appendix \ref{appendix:prompt}.

\paragraph{Hyperparameters.}
During evaluation, we restrict the maximum sequence length to 4,096 tokens and set the temperature parameter to 0.0, except for Multi-News+, which uses a sequence length of 8,192 tokens.
We disable the thinking mode for Qwen3.6 and Nemotron3 and assign a low reasoning effort setting for GPT-OSS by default.\looseness=-1

For quantization, we employ a W4A16 quantization scheme (4-bit weights and 16-bit activations) targeting all linear layers in the model. The language modeling head (\texttt{lm\_head}) and the MoE gate/router layers (\texttt{gate} and \texttt{router} parameters) are excluded from quantization and kept in bfloat16 precision. For calibration, we use the same calibration set as in pruning (i.e., 128 randomly sampled instances from MedINST).

\paragraph{Human Evaluation Protocol.}
To validate the LLM-as-a-Judge evaluation framework, we conduct a manual annotation study to compare LLM judgments with human annotations. 
We recruit three PhD student volunteers with experience in natural language processing and computer science.
Because the evaluation protocol relies strictly on textual grounding, i.e., verifying whether statements in the generated summary are logically entailed by the provided source documents, rather than requiring external clinical diagnostic knowledge, these annotators are highly qualified to evaluate semantic alignment and omission errors.
The human annotators are provided with the same instructions and prompt guidelines as the LLM judges (documented in Appendix~\ref{appendix:prompt}).
Specifically, for absolute judgment, we ask the annotators to classify the summaries generated by the pruned models as either faithful or hallucinated. For relative judgment, they compare the summaries from the unpruned baseline against the pruned models across four dimensions (hallucinations, omissions, repetition, and semantic alignment). Both evaluations are conducted using only the provided source documents.\looseness=-1

The human evaluation is conducted on a subset of 30 randomly selected documents from the RCT and Multi-XScience datasets. The sample distribution of these 30 evaluation cases across datasets, generator models, pruning methods, and pruning ratios is structured as follows:
\begin{itemize}
    \item \textbf{Dataset Balance}: 15 samples from RCT and 15 samples from Multi-XScience.
    \item \textbf{Generator Model Balance}:
    \begin{itemize}
        \item \textbf{GPT-OSS}: 8 samples (4 RCT, 4 Multi-XScience)
        \item \textbf{Qwen3}: 7 samples (4 RCT, 3 Multi-XScience)
        \item \textbf{Qwen3.6}: 7 samples (3 RCT, 4 Multi-XScience)
        \item \textbf{Nemotron3}: 8 samples (4 RCT, 4 Multi-XScience)
    \end{itemize}
    \item \textbf{Pruning Method Balance}: 5 samples per pruning method, distributed evenly across all six approaches (Random, Frequency, Gate, EAN, EASY-EP, and REAP).
    \item \textbf{Pruning Ratio Balance}: 
    \begin{itemize}
        \item Ratio 0.125: 4 samples
        \item Ratio 0.250: 4 samples
        \item Ratio 0.375: 5 samples
        \item Ratio 0.500: 4 samples
        \item Ratio 0.625: 4 samples
        \item Ratio 0.750: 5 samples
        \item Ratio 0.875: 4 samples
    \end{itemize}
\end{itemize}
All pruning ratios within each individual (dataset, generator model, pruning method) group are guaranteed to be distinct for the selected samples to ensure a representative and unbiased distribution across the entire pruning range.

\section{Prompt Templates}
\label{appendix:prompt}

\begin{examplebox}{RCT prompt template}
You are a medical evidence summarization assistant. Given a randomized controlled trial abstract, write a concise factual summary in 2-4 sentences. Do not include information that is not supported by the abstract.\\

Source documents:
\{source\}\\

Summary:
\end{examplebox}

\begin{examplebox}{Multi-News+ prompt template}
You are a news summarization assistant. Given documents from multiple news sources about the same event, write a concise factual summary in 2-4 sentences. Do not include information that is not supported by the source.\\

Source documents:
\{source\}\\

Summary:
\end{examplebox}

\begin{examplebox}{Absolute judgment prompt template}
Given a list of factual alignments and contradictions, which highlights alignment/contradictions between the `actual output' and `contexts', use it to provide a reason for the hallucination score CONCISELY. Note that the hallucination score ranges from 0 - 1, and the lower the better.\\

** 
IMPORTANT: Please make sure to only return in JSON format, with the `reason' key providing the reason.
Example JSON:
\{\{
    "reason": "The score is <hallucination\_score> because <your\_reason>."
\}\}
**\\

Factual Alignments:
\{factual\_alignments\}\\

Contradictions:
\{contradictions\}\\

Hallucination Score:
\{score\}\\

JSON:
\end{examplebox}

\begin{examplebox}{Relative judgment prompt template}
You are an impartial evaluator. Compare Summary A vs Summary B against the Source Document.
Use ONLY the Source Document to judge support. Do NOT use outside knowledge.
Be strict about factual support.\\

Answer these questions:\\

Q1. Hallucinations: Which summary contains MORE hallucinations (unsupported content)?
Q2. Omission: Which summary is missing MORE crucial information from the document?
Q3. Repetition: Which summary contains MORE repetitive information?
Q4. Alignment: Which summary is MORE semantically aligned with the source document?\\

Return ONLY valid JSON with exactly these keys:
- q1\_hallucinations\_more: "A" or "B"
- q2\_omission\_more: "A" or "B"
- q3\_repetition\_more: "A" or "B"
- q4\_alignment\_more: "A" or "B"\\

Rules:
- You MUST choose `A' or `B' (no ties).
- Keep outputs to JSON only (no markdown).\\

Source Document:
<<<DOC
\{document\}
DOC>>>\\

Summary A:
<<<A
\{summary\_a\}
A>>>\\

Summary B:
<<<B
\{summary\_b\}
B>>>
\end{examplebox}

\clearpage
\section{Supplementary Results}
\input{table/annotator_agreement}
\begin{figure}
    \centering
    \includegraphics[width=\linewidth]{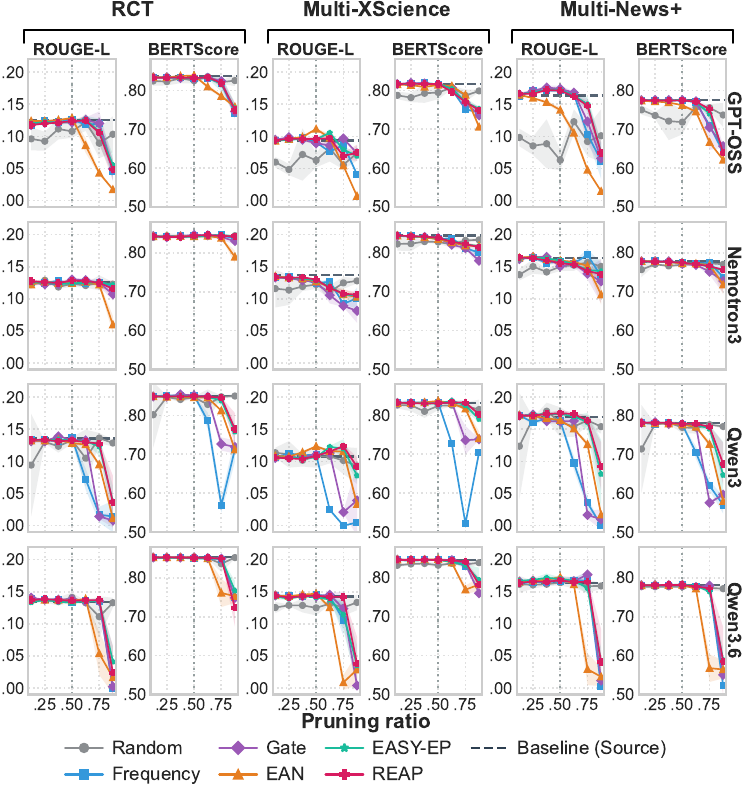}
    \caption{Summarization performance across different pruning ratios and approaches.}
    \label{fig:sum_ratio_results}
\end{figure}
\input{table/utility_general}
\input{table/utility_gen_target}
\input{table/utility_cls_target}

\end{document}

%% file: table/human_llm_agreement.tex
\begin{table}[t]
\centering
\renewcommand{\arraystretch}{0.8}
\setlength{\tabcolsep}{3pt}
\setlength{\aboverulesep}{1.3pt}
\setlength{\belowrulesep}{1.3pt}
\resizebox{0.9\linewidth}{!}{
\begin{tabular}{ll ccc}
\toprule
 & \textbf{Pair} & \textbf{Low} & \textbf{Moderate} & \textbf{Extreme} \\
\midrule
\multicolumn{2}{l}{\textbf{Sample Size ($N$)}} & 8 & 9 & 13 \\
\midrule
\multirow{3}{*}{\textbf{Absolute}} 
 & Human--Human & 0.467 & 1.000 & 0.898 \\
 & Human--LLM   & 0.006 & 0.738 & 0.644 \\
 & LLM--LLM     & 0.333 & 0.738 & 0.783 \\
\midrule
\multirow{3}{*}{\textbf{Relative}} 
 & Human--Human & 0.366 & 0.260 & 0.733 \\
 & Human--LLM   & 0.238 & -0.012 & 0.605 \\
 & LLM--LLM     & 0.508 & 0.370 & 0.676 \\
\bottomrule
\end{tabular}
}
\caption{Reliability agreement (Cohen's $\kappa$) across expert pruning ratio windows. $\kappa$ is computed pairwise for all pairs within a category (3 pairs for Human--Human and LLM--LLM, and 9 pairs for Human--LLM) and then averaged. Relative agreement is pooled across questions (Q1--Q4) for each pair before computing the metrics.}

\label{tab:agreement_by_window}
\end{table}

%% file: table/ablation.tex
\begin{table*}[t]
\centering
\renewcommand{\arraystretch}{0.8}
\setlength{\tabcolsep}{3pt}
\setlength{\aboverulesep}{1.3pt}
\setlength{\belowrulesep}{1.3pt}
\resizebox{\textwidth}{!}{
\begin{tabular}{l ccc cccccc ccccc}
\toprule
\textbf{Configuration} & \multicolumn{3}{c}{\textbf{Utility}} & \multicolumn{6}{c}{\textbf{Biomedical Reliability}} & \multicolumn{5}{c}{\textbf{General Reliability}} \\
\cmidrule(lr){2-4} \cmidrule(lr){5-10} \cmidrule(lr){11-15}
 & \textbf{MedINST} & \textbf{MMQA} & \textbf{General} & \textbf{RCT} & \textbf{MXSci.} & \textbf{Hall.} & \textbf{Omis.} & \textbf{Repe.} & \textbf{Align.} & \textbf{MN+} & \textbf{Hall.} & \textbf{Omis.} & \textbf{Repe.} & \textbf{Align.} \\
\midrule
MedINST (50\%) & \textbf{.554}\textsubscript{.004} & \textbf{.838}\textsubscript{.008} & .689\textsubscript{.010} & \textbf{.015}\textsubscript{.009} & .121\textsubscript{.079} & .476\textsubscript{.029} & \textbf{.402}\textsubscript{.073} & \textbf{.516}\textsubscript{.067} & \textbf{.564}\textsubscript{.036} & \textbf{.031}\textsubscript{.036} & \textbf{.367}\textsubscript{.076} & \textbf{.480}\textsubscript{.106} & \textbf{.547}\textsubscript{.133} & \textbf{.647}\textsubscript{.012} \\
+ General (50\%) & .548\textsubscript{.013} & .833\textsubscript{.004} & \textbf{.915}\textsubscript{.064} & .019\textsubscript{.017} & \textbf{.088}\textsubscript{.064} & \textbf{.437}\textsubscript{.033} & .470\textsubscript{.044} & .517\textsubscript{.053} & .550\textsubscript{.029} & .060\textsubscript{.045} & .510\textsubscript{.082} & .537\textsubscript{.065} & .593\textsubscript{.094} & .447\textsubscript{.079} \\
\midrule
MedINST (75\%) & \textbf{.476}\textsubscript{.037} & \textbf{.665}\textsubscript{.137} & .387\textsubscript{.088} & \textbf{.024}\textsubscript{.019} & \textbf{.179}\textsubscript{.044} & \textbf{.204}\textsubscript{.029} & .735\textsubscript{.088} & \textbf{.405}\textsubscript{.161} & \textbf{.417}\textsubscript{.079} & \textbf{.211}\textsubscript{.073} & \textbf{.540}\textsubscript{.020} & .727\textsubscript{.081} & \textbf{.680}\textsubscript{.140} & \textbf{.287}\textsubscript{.031} \\
+ General (75\%) & .424\textsubscript{.128} & .563\textsubscript{.096} & \textbf{.597}\textsubscript{.287} & .061\textsubscript{.026} & .358\textsubscript{.025} & .372\textsubscript{.050} & \textbf{.646}\textsubscript{.040} & .653\textsubscript{.112} & .414\textsubscript{.029} & .323\textsubscript{.167} & .755\textsubscript{.100} & \textbf{.630}\textsubscript{.162} & .905\textsubscript{.055} & .160\textsubscript{.043} \\
\midrule
+ GPTQ (4-bit) & .543\textsubscript{.005} & .829\textsubscript{.011} & .676\textsubscript{.020} & .017\textsubscript{.026} & .125\textsubscript{.079} & \textbf{.419}\textsubscript{.025} & .470\textsubscript{.037} & \textbf{.492}\textsubscript{.022} & .564\textsubscript{.034} & .078\textsubscript{.055} & .547\textsubscript{.058} & .553\textsubscript{.085} & .620\textsubscript{.078} & .410\textsubscript{.075} \\
\bottomrule
\end{tabular}
}
\caption{Utility and reliability metrics for Qwen3.6 EASY-EP pruned models, contrasting biomedical-only calibration against dual-domain calibration at 50\% and 75\% ratios, alongside the impact of 4-bit GPTQ quantization. Bold text denotes superior performance within each pairing; subscripts indicate standard deviation across three seeds.}
\label{tab:ablation_results}
\end{table*}

%% file: table/annotator_agreement.tex
\begin{table}[h]
\centering
\small
\begin{tabular}{lc}
\toprule
\textbf{Dataset} & \textbf{Overall Fleiss' $\kappa$} \\
\midrule
Multi-News & .577 \\
Multi-XScience & .513 \\
RCT & .535 \\
\bottomrule
\end{tabular}
\caption{Inter-annotator agreement (Fleiss' $\kappa$) for absolute judgments among three LLMs.}
\label{tab:agreement_absolute}
\end{table}

\begin{table}[h]
\centering
\small
\resizebox{\linewidth}{!}{
\begin{tabular}{llc}
\toprule
\textbf{Dataset} & \textbf{Question} & \textbf{Overall Fleiss' $\kappa$} \\
\midrule
\multirow{4}{*}{Multi-News} 
 & Q1: Hallucination & .503 \\
 & Q2: Omission & .447 \\
 & Q3: Repetition & .285 \\
 & Q4: Alignment & .519 \\
\midrule
\multirow{4}{*}{Multi-XScience} 
 & Q1: Hallucination & .484 \\
 & Q2: Omission & .336 \\
 & Q3: Repetition & .456 \\
 & Q4: Alignment & .555 \\
\midrule
\multirow{4}{*}{RCT} 
 & Q1: Hallucination & .483 \\
 & Q2: Omission & .492 \\
 & Q3: Repetition & .370 \\
 & Q4: Alignment & .470 \\
\bottomrule
\end{tabular}
}
\caption{Inter-annotator agreement (Fleiss' $\kappa$) for relative judgments among three LLMs.}
\label{tab:agreement_relative}
\end{table}

%% file: table/utility_general.tex
\begin{table}[t]
\centering
\small
\resizebox{\linewidth}{!}{
\begin{tabular}{clcccc}
\toprule
& \textbf{Approach} & \textbf{IFEval} & \textbf{GSM8K} & \textbf{HumanEval} & \textbf{MMLU} \\
\midrule
\multirow{7}{*}{\rotatebox{90}{GPT-OSS}} & \cellcolor{gray!20}Source & \cellcolor{gray!20}.750 & \cellcolor{gray!20}.056 & \cellcolor{gray!20}.817 & \cellcolor{gray!20}.843 \\ \cmidrule{2-6}
 & Random & .221\textsubscript{.008} & \textbf{.009\textsubscript{.015}} & \textbf{.163\textsubscript{.135}} & .265\textsubscript{.130} \\
 & Frequency & \textbf{.267\textsubscript{.001}} & .001\textsubscript{.001} & .018\textsubscript{.009} & \underline{.656\textsubscript{.006}} \\
 & Gate & .251\textsubscript{.010} & \underline{.007\textsubscript{.003}} & .012\textsubscript{.000} & \textbf{.660\textsubscript{.003}} \\
 & EAN & .252\textsubscript{.003} & .001\textsubscript{.002} & \underline{.073\textsubscript{.006}} & .534\textsubscript{.015} \\
 & EASY-EP & \underline{.261\textsubscript{.007}} & .003\textsubscript{.003} & .014\textsubscript{.004} & .655\textsubscript{.006} \\
 & REAP & .260\textsubscript{.004} & .005\textsubscript{.002} & .012\textsubscript{.000} & \underline{.656\textsubscript{.002}} \\
\midrule
\multirow{7}{*}{\rotatebox{90}{Nemotron3}} & \cellcolor{gray!20}Source & \cellcolor{gray!20}.813 & \cellcolor{gray!20}.879 & \cellcolor{gray!20}.848 & \cellcolor{gray!20}.723 \\ \cmidrule{2-6}
 & Random & .555\textsubscript{.034} & .378\textsubscript{.040} & \underline{.053\textsubscript{.039}} & .497\textsubscript{.021} \\
 & Frequency & .668\textsubscript{.019} & .148\textsubscript{.011} & .006\textsubscript{.000} & .545\textsubscript{.001} \\
 & Gate & .683\textsubscript{.005} & .159\textsubscript{.007} & .010\textsubscript{.004} & \underline{.552\textsubscript{.008}} \\
 & EAN & .653\textsubscript{.015} & .372\textsubscript{.031} & \textbf{.254\textsubscript{.029}} & .526\textsubscript{.003} \\
 & EASY-EP & \textbf{.702\textsubscript{.015}} & \textbf{.472\textsubscript{.022}} & .012\textsubscript{.006} & \textbf{.565\textsubscript{.007}} \\
 & REAP & \underline{.689\textsubscript{.008}} & \underline{.441\textsubscript{.028}} & .006\textsubscript{.000} & \textbf{.565\textsubscript{.005}} \\
\midrule
\multirow{7}{*}{\rotatebox{90}{Qwen3}} & \cellcolor{gray!20}Source & \cellcolor{gray!20}.806 & \cellcolor{gray!20}.835 & \cellcolor{gray!20}.939 & \cellcolor{gray!20}.801 \\ \cmidrule{2-6}
 & Random & .424\textsubscript{.100} & .320\textsubscript{.198} & \underline{.021\textsubscript{.004}} & .437\textsubscript{.126} \\
 & Frequency & .578\textsubscript{.015} & .547\textsubscript{.034} & .000\textsubscript{.000} & .569\textsubscript{.013} \\
 & Gate & .663\textsubscript{.016} & .573\textsubscript{.066} & .000\textsubscript{.000} & .585\textsubscript{.005} \\
 & EAN & \underline{.717\textsubscript{.008}} & \textbf{\cellcolor{green!20}.836\textsubscript{.011}} & \textbf{.882\textsubscript{.015}} & .559\textsubscript{.002} \\
 & EASY-EP & \textbf{.779\textsubscript{.012}} & .692\textsubscript{.008} & .014\textsubscript{.004} & \textbf{.668\textsubscript{.005}} \\
 & REAP & \textbf{.779\textsubscript{.003}} & \underline{.695\textsubscript{.006}} & .008\textsubscript{.004} & \underline{.656\textsubscript{.003}} \\
\midrule
\multirow{7}{*}{\rotatebox{90}{Qwen3.6}} & \cellcolor{gray!20}Source & \cellcolor{gray!20}.824 & \cellcolor{gray!20}.880 & \cellcolor{gray!20}.970 & \cellcolor{gray!20}.814 \\ \cmidrule{2-6}
 & Random & .581\textsubscript{.084} & .178\textsubscript{.200} & \underline{.217\textsubscript{.057}} & .530\textsubscript{.104} \\
 & Frequency & .770\textsubscript{.004} & .106\textsubscript{.047} & .018\textsubscript{.006} & .644\textsubscript{.020} \\
 & Gate & .773\textsubscript{.011} & .128\textsubscript{.044} & .008\textsubscript{.004} & .642\textsubscript{.006} \\
 & EAN & \textbf{.776\textsubscript{.021}} & \textbf{.862\textsubscript{.010}} & \textbf{.780\textsubscript{.012}} & \underline{.680\textsubscript{.013}} \\
 & EASY-EP & .746\textsubscript{.015} & \underline{.778\textsubscript{.027}} & .083\textsubscript{.009} & \textbf{.719\textsubscript{.008}} \\
 & REAP & \underline{.774\textsubscript{.004}} & .766\textsubscript{.008} & .039\textsubscript{.009} & .663\textsubscript{.003} \\
\bottomrule
\end{tabular}
}
\caption{Downstream performance on general-domain benchmarks (IFEval, GSM8K, HumanEval, and MMLU) at 50\% expert pruning ratio. Scores that are better than Source are highlighted in \colorbox{green!20}{green}. The best and second-best approaches for each model family are in \textbf{bold} and \underline{underlined}, respectively. Subscripts denote standard deviation across three seeds.}
\label{tab:utility_general}
\end{table}

%% file: table/utility_gen_target.tex
\begin{table*}[t]
\centering
\small
\resizebox{\textwidth}{!}{
\begin{tabular}{llcccccccc}
\toprule
 & \textbf{Approach}  & \textbf{SUM}  & \textbf{MT}  & \textbf{QA}  & \textbf{NER}  & \textbf{NED}  & \textbf{RE}  & \textbf{COREF}  & \textbf{EE}  \\
\midrule
\multirow{7}{*}{\rotatebox{90}{GPT-OSS}} & \cellcolor{gray!20}Source & \cellcolor{gray!20}.100 & \cellcolor{gray!20}.585 & \cellcolor{gray!20}.768 & \cellcolor{gray!20}.538 & \cellcolor{gray!20}.331 & \cellcolor{gray!20}.342 & \cellcolor{gray!20}.419 & \cellcolor{gray!20}.290 \\ \cmidrule{2-10}
 & Random & .066\textsubscript{.011} & .078\textsubscript{.042} & .216\textsubscript{.255} & .124\textsubscript{.025} & .035\textsubscript{.005} & .089\textsubscript{.059} & .008\textsubscript{.006} & .088\textsubscript{.039} \\
 & Frequency & .097\textsubscript{.000} & \textbf{.536\textsubscript{.004}} & .715\textsubscript{.008} & .505\textsubscript{.007} & .309\textsubscript{.009} & \cellcolor{green!20}.345\textsubscript{.024} & \textbf{\cellcolor{green!20}.521\textsubscript{.060}} & .274\textsubscript{.017} \\
 & Gate & .096\textsubscript{.002} & \underline{.531\textsubscript{.011}} & \textbf{.738\textsubscript{.005}} & .507\textsubscript{.005} & \underline{.313\textsubscript{.001}} & \textbf{\cellcolor{green!20}.358\textsubscript{.007}} & \underline{\cellcolor{green!20}.492\textsubscript{.099}} & \textbf{.283\textsubscript{.006}} \\
 & EAN & \textbf{\cellcolor{green!20}.119\textsubscript{.001}} & .285\textsubscript{.022} & .636\textsubscript{.003} & .331\textsubscript{.007} & .162\textsubscript{.000} & \underline{\cellcolor{green!20}.351\textsubscript{.010}} & .368\textsubscript{.044} & .138\textsubscript{.005} \\
 & EASY-EP & \cellcolor{green!20}.102\textsubscript{.003} & .498\textsubscript{.028} & .722\textsubscript{.009} & \textbf{.517\textsubscript{.006}} & \textbf{.317\textsubscript{.009}} & .316\textsubscript{.030} & \cellcolor{green!20}.447\textsubscript{.045} & \underline{.280\textsubscript{.016}} \\
 & REAP & \underline{\cellcolor{green!20}.103\textsubscript{.004}} & .485\textsubscript{.018} & \underline{.726\textsubscript{.006}} & \underline{.516\textsubscript{.005}} & \textbf{.317\textsubscript{.019}} & .317\textsubscript{.033} & \cellcolor{green!20}.459\textsubscript{.050} & \textbf{.283\textsubscript{.015}} \\
\midrule
\multirow{7}{*}{\rotatebox{90}{Nemotron3}} & \cellcolor{gray!20}Source & \cellcolor{gray!20}.148 & \cellcolor{gray!20}.515 & \cellcolor{gray!20}.735 & \cellcolor{gray!20}.445 & \cellcolor{gray!20}.334 & \cellcolor{gray!20}.384 & \cellcolor{gray!20}.695 & \cellcolor{gray!20}.244 \\ \cmidrule{2-10}
 & Random & .130\textsubscript{.007} & .319\textsubscript{.013} & .580\textsubscript{.068} & .355\textsubscript{.009} & .183\textsubscript{.007} & \underline{.370\textsubscript{.016}} & .648\textsubscript{.020} & .158\textsubscript{.031} \\
 & Frequency & .132\textsubscript{.001} & .305\textsubscript{.037} & \textbf{.692\textsubscript{.005}} & .431\textsubscript{.003} & .276\textsubscript{.004} & .354\textsubscript{.006} & \cellcolor{green!20}.695\textsubscript{.045} & .199\textsubscript{.004} \\
 & Gate & .136\textsubscript{.000} & .375\textsubscript{.015} & \underline{.686\textsubscript{.007}} & .434\textsubscript{.001} & .286\textsubscript{.004} & .366\textsubscript{.008} & \cellcolor{green!20}.698\textsubscript{.077} & .228\textsubscript{.016} \\
 & EAN & .135\textsubscript{.002} & \textbf{.441\textsubscript{.017}} & .676\textsubscript{.008} & .422\textsubscript{.003} & .266\textsubscript{.004} & .353\textsubscript{.012} & \cellcolor{green!20}.714\textsubscript{.050} & \underline{.235\textsubscript{.010}} \\
 & EASY-EP & \underline{.137\textsubscript{.003}} & .365\textsubscript{.064} & .682\textsubscript{.006} & \textbf{.439\textsubscript{.002}} & \textbf{.298\textsubscript{.002}} & .356\textsubscript{.007} & \textbf{\cellcolor{green!20}.757\textsubscript{.012}} & .215\textsubscript{.002} \\
 & REAP & \textbf{.139\textsubscript{.001}} & \underline{.435\textsubscript{.052}} & .684\textsubscript{.015} & \underline{.438\textsubscript{.002}} & \underline{.296\textsubscript{.003}} & \textbf{.382\textsubscript{.015}} & \underline{\cellcolor{green!20}.751\textsubscript{.015}} & \textbf{.236\textsubscript{.003}} \\
\midrule
\multirow{7}{*}{\rotatebox{90}{Qwen3}} & \cellcolor{gray!20}Source & \cellcolor{gray!20}.115 & \cellcolor{gray!20}.615 & \cellcolor{gray!20}.663 & \cellcolor{gray!20}.566 & \cellcolor{gray!20}.407 & \cellcolor{gray!20}.396 & \cellcolor{gray!20}.802 & \cellcolor{gray!20}.368 \\ \cmidrule{2-10}
 & Random & .114\textsubscript{.019} & .371\textsubscript{.054} & .603\textsubscript{.035} & .366\textsubscript{.033} & .154\textsubscript{.031} & .366\textsubscript{.030} & .729\textsubscript{.049} & .202\textsubscript{.048} \\
 & Frequency & \cellcolor{green!20}.118\textsubscript{.002} & .450\textsubscript{.024} & .597\textsubscript{.018} & .529\textsubscript{.013} & .347\textsubscript{.010} & \underline{.393\textsubscript{.025}} & .747\textsubscript{.038} & .336\textsubscript{.018} \\
 & Gate & \underline{\cellcolor{green!20}.119\textsubscript{.001}} & .484\textsubscript{.003} & \cellcolor{green!20}.665\textsubscript{.025} & .545\textsubscript{.001} & .355\textsubscript{.012} & \textbf{.394\textsubscript{.014}} & \textbf{\cellcolor{green!20}.825\textsubscript{.003}} & .360\textsubscript{.003} \\
 & EAN & \textbf{\cellcolor{green!20}.132\textsubscript{.002}} & .442\textsubscript{.019} & \cellcolor{green!20}.663\textsubscript{.007} & .555\textsubscript{.002} & .271\textsubscript{.004} & .370\textsubscript{.006} & .793\textsubscript{.016} & .355\textsubscript{.006} \\
 & EASY-EP & .114\textsubscript{.002} & \textbf{.499\textsubscript{.003}} & \underline{\cellcolor{green!20}.752\textsubscript{.006}} & \textbf{\cellcolor{green!20}.575\textsubscript{.003}} & \textbf{\cellcolor{green!20}.410\textsubscript{.001}} & .382\textsubscript{.020} & \underline{\cellcolor{green!20}.823\textsubscript{.014}} & \underline{.368\textsubscript{.003}} \\
 & REAP & \cellcolor{green!20}.116\textsubscript{.002} & \underline{.488\textsubscript{.004}} & \textbf{\cellcolor{green!20}.753\textsubscript{.002}} & \underline{\cellcolor{green!20}.571\textsubscript{.001}} & \underline{.402\textsubscript{.006}} & .377\textsubscript{.008} & .794\textsubscript{.017} & \textbf{\cellcolor{green!20}.374\textsubscript{.005}} \\
\midrule
\multirow{7}{*}{\rotatebox{90}{Qwen3.6}} & \cellcolor{gray!20}Source & \cellcolor{gray!20}.154 & \cellcolor{gray!20}.624 & \cellcolor{gray!20}.566 & \cellcolor{gray!20}.714 & \cellcolor{gray!20}.427 & \cellcolor{gray!20}.479 & \cellcolor{gray!20}.899 & \cellcolor{gray!20}.304 \\ \cmidrule{2-10}
 & Random & .133\textsubscript{.012} & .540\textsubscript{.009} & \cellcolor{green!20}.614\textsubscript{.066} & .519\textsubscript{.016} & .204\textsubscript{.038} & .430\textsubscript{.021} & .764\textsubscript{.099} & .218\textsubscript{.065} \\
 & Frequency & \underline{\cellcolor{green!20}.157\textsubscript{.001}} & \textbf{.617\textsubscript{.003}} & \cellcolor{green!20}.760\textsubscript{.015} & .702\textsubscript{.006} & .407\textsubscript{.012} & .475\textsubscript{.018} & \underline{\cellcolor{green!20}.926\textsubscript{.018}} & \cellcolor{green!20}.323\textsubscript{.007} \\
 & Gate & .154\textsubscript{.002} & \textbf{.617\textsubscript{.002}} & \cellcolor{green!20}.653\textsubscript{.048} & \underline{.704\textsubscript{.003}} & .423\textsubscript{.004} & \underline{\cellcolor{green!20}.494\textsubscript{.007}} & \textbf{\cellcolor{green!20}.928\textsubscript{.018}} & \cellcolor{green!20}.322\textsubscript{.004} \\
 & EAN & \textbf{\cellcolor{green!20}.158\textsubscript{.003}} & .605\textsubscript{.009} & \underline{\cellcolor{green!20}.795\textsubscript{.005}} & .690\textsubscript{.007} & .366\textsubscript{.009} & .449\textsubscript{.002} & \cellcolor{green!20}.907\textsubscript{.004} & .286\textsubscript{.019} \\
 & EASY-EP & .152\textsubscript{.003} & \underline{.612\textsubscript{.007}} & \textbf{\cellcolor{green!20}.797\textsubscript{.011}} & .700\textsubscript{.018} & \underline{\cellcolor{green!20}.427\textsubscript{.018}} & \cellcolor{green!20}.480\textsubscript{.017} & \textbf{\cellcolor{green!20}.928\textsubscript{.014}} & \underline{\cellcolor{green!20}.333\textsubscript{.010}} \\
 & REAP & \cellcolor{green!20}.154\textsubscript{.003} & \textbf{.617\textsubscript{.004}} & \cellcolor{green!20}.595\textsubscript{.032} & \textbf{.710\textsubscript{.002}} & \textbf{\cellcolor{green!20}.437\textsubscript{.002}} & \textbf{\cellcolor{green!20}.499\textsubscript{.009}} & .887\textsubscript{.011} & \textbf{\cellcolor{green!20}.341\textsubscript{.009}} \\
\bottomrule
\end{tabular}
}
\caption{Zero-shot performance on unseen MedINST downstream tasks. Scores that are better than Source are highlighted in \colorbox{green!20}{green}. The best and second-best adaptation approaches for each model family are indicated in \textbf{bold} and \underline{underlined}, respectively. Subscripts denote standard deviation across three seeds.}
\label{tab:utility_gen_target}
\end{table*}

%% file: table/utility_cls_target.tex
\begin{table*}[t]
\centering
\small
\resizebox{\textwidth}{!}{
\begin{tabular}{clccccccccc}
\toprule
& \textbf{Approach} & \textbf{PMQA} & \textbf{MQA4} & \textbf{MMCQA} & \textbf{Anat} & \textbf{CK} & \textbf{CM} & \textbf{MG} & \textbf{PM} & \textbf{CB} \\
\midrule
\multirow{7}{*}{\rotatebox{90}{GPT-OSS}} & \cellcolor{gray!20}Source & \cellcolor{gray!20}.688 & \cellcolor{gray!20}.791 & \cellcolor{gray!20}.653 & \cellcolor{gray!20}.793 & \cellcolor{gray!20}.834 & \cellcolor{gray!20}.850 & \cellcolor{gray!20}.940 & \cellcolor{gray!20}.915 & \cellcolor{gray!20}.944 \\ \cmidrule{2-11}
 & Random & .347\textsubscript{.260} & .081\textsubscript{.042} & .140\textsubscript{.089} & .232\textsubscript{.137} & .260\textsubscript{.168} & .229\textsubscript{.110} & .233\textsubscript{.150} & .114\textsubscript{.042} & .319\textsubscript{.211} \\
 & Frequency & .667\textsubscript{.011} & .719\textsubscript{.014} & \underline{.568\textsubscript{.004}} & \textbf{.731\textsubscript{.033}} & .792\textsubscript{.004} & \underline{.721\textsubscript{.027}} & .807\textsubscript{.015} & \underline{.871\textsubscript{.007}} & .829\textsubscript{.041} \\
 & Gate & .682\textsubscript{.024} & \textbf{.732\textsubscript{.012}} & \textbf{.569\textsubscript{.003}} & .689\textsubscript{.020} & \underline{.794\textsubscript{.029}} & .707\textsubscript{.018} & \textbf{.827\textsubscript{.045}} & \textbf{.881\textsubscript{.024}} & .808\textsubscript{.008} \\
 & EAN & \cellcolor{green!20}.713\textsubscript{.004} & .312\textsubscript{.011} & .337\textsubscript{.002} & .363\textsubscript{.013} & .499\textsubscript{.008} & .459\textsubscript{.022} & .417\textsubscript{.038} & .391\textsubscript{.009} & .521\textsubscript{.032} \\
 & EASY-EP & \textbf{\cellcolor{green!20}.722\textsubscript{.016}} & \underline{.721\textsubscript{.005}} & .553\textsubscript{.004} & .694\textsubscript{.031} & \textbf{.801\textsubscript{.014}} & \textbf{.730\textsubscript{.003}} & \underline{.813\textsubscript{.023}} & .864\textsubscript{.025} & \textbf{.838\textsubscript{.016}} \\
 & REAP & \underline{\cellcolor{green!20}.717\textsubscript{.018}} & .705\textsubscript{.006} & .552\textsubscript{.004} & \underline{.719\textsubscript{.026}} & .777\textsubscript{.021} & \textbf{.730\textsubscript{.007}} & \underline{.813\textsubscript{.021}} & .864\textsubscript{.004} & \underline{.833\textsubscript{.014}} \\
\midrule
\multirow{7}{*}{\rotatebox{90}{Nemotron3}} & \cellcolor{gray!20}Source & \cellcolor{gray!20}.440 & \cellcolor{gray!20}.636 & \cellcolor{gray!20}.545 & \cellcolor{gray!20}.704 & \cellcolor{gray!20}.770 & \cellcolor{gray!20}.717 & \cellcolor{gray!20}.840 & \cellcolor{gray!20}.831 & \cellcolor{gray!20}.875 \\ \cmidrule{2-11}
 & Random & \underline{\cellcolor{green!20}.479\textsubscript{.021}} & .364\textsubscript{.066} & .348\textsubscript{.042} & .410\textsubscript{.009} & .513\textsubscript{.047} & .499\textsubscript{.023} & .507\textsubscript{.021} & .506\textsubscript{.024} & .507\textsubscript{.025} \\
 & Frequency & .381\textsubscript{.041} & .510\textsubscript{.010} & .436\textsubscript{.008} & .580\textsubscript{.026} & .657\textsubscript{.023} & .640\textsubscript{.009} & \textbf{.767\textsubscript{.023}} & .713\textsubscript{.007} & .653\textsubscript{.021} \\
 & Gate & .398\textsubscript{.026} & .502\textsubscript{.021} & .449\textsubscript{.005} & .595\textsubscript{.011} & .669\textsubscript{.034} & .651\textsubscript{.033} & .737\textsubscript{.021} & .729\textsubscript{.006} & .671\textsubscript{.011} \\
 & EAN & \textbf{\cellcolor{green!20}.609\textsubscript{.021}} & .434\textsubscript{.006} & .404\textsubscript{.012} & .479\textsubscript{.041} & .575\textsubscript{.014} & .559\textsubscript{.009} & .723\textsubscript{.025} & .614\textsubscript{.017} & .667\textsubscript{.025} \\
 & EASY-EP & .397\textsubscript{.012} & \textbf{.532\textsubscript{.007}} & \textbf{.472\textsubscript{.006}} & \textbf{.607\textsubscript{.022}} & \textbf{.709\textsubscript{.016}} & \textbf{.682\textsubscript{.015}} & \underline{.747\textsubscript{.031}} & \textbf{.745\textsubscript{.014}} & \textbf{.713\textsubscript{.004}} \\
 & REAP & .427\textsubscript{.031} & \underline{.520\textsubscript{.004}} & \underline{.471\textsubscript{.003}} & \underline{.605\textsubscript{.004}} & \underline{.702\textsubscript{.007}} & \underline{.678\textsubscript{.009}} & .740\textsubscript{.026} & \underline{.733\textsubscript{.011}} & \underline{.706\textsubscript{.034}} \\
\midrule
\multirow{7}{*}{\rotatebox{90}{Qwen3}} & \cellcolor{gray!20}Source & \cellcolor{gray!20}.700 & \cellcolor{gray!20}.761 & \cellcolor{gray!20}.670 & \cellcolor{gray!20}.778 & \cellcolor{gray!20}.879 & \cellcolor{gray!20}.821 & \cellcolor{gray!20}.930 & \cellcolor{gray!20}.890 & \cellcolor{gray!20}.924 \\ \cmidrule{2-11}
 & Random & .599\textsubscript{.045} & .359\textsubscript{.078} & .360\textsubscript{.032} & .378\textsubscript{.181} & .454\textsubscript{.132} & .432\textsubscript{.139} & .420\textsubscript{.075} & .415\textsubscript{.160} & .463\textsubscript{.149} \\
 & Frequency & .661\textsubscript{.029} & .645\textsubscript{.012} & .553\textsubscript{.008} & .709\textsubscript{.024} & .779\textsubscript{.019} & .701\textsubscript{.009} & .817\textsubscript{.015} & .786\textsubscript{.018} & .889\textsubscript{.007} \\
 & Gate & .671\textsubscript{.013} & .667\textsubscript{.017} & .565\textsubscript{.014} & .728\textsubscript{.017} & .784\textsubscript{.023} & .734\textsubscript{.010} & .817\textsubscript{.015} & .821\textsubscript{.024} & .870\textsubscript{.011} \\
 & EAN & .633\textsubscript{.008} & .432\textsubscript{.005} & .398\textsubscript{.004} & .393\textsubscript{.013} & .591\textsubscript{.023} & .563\textsubscript{.023} & .593\textsubscript{.015} & .569\textsubscript{.017} & .632\textsubscript{.000} \\
 & EASY-EP & \underline{.695\textsubscript{.010}} & \textbf{.744\textsubscript{.007}} & \textbf{.646\textsubscript{.002}} & \textbf{.773\textsubscript{.030}} & \textbf{.835\textsubscript{.004}} & \underline{.782\textsubscript{.007}} & \underline{.870\textsubscript{.010}} & \textbf{.880\textsubscript{.002}} & \underline{.917\textsubscript{.000}} \\
 & REAP & \textbf{.699\textsubscript{.010}} & \underline{.737\textsubscript{.003}} & \underline{.640\textsubscript{.002}} & \underline{.770\textsubscript{.020}} & \underline{.830\textsubscript{.008}} & \textbf{.786\textsubscript{.006}} & \textbf{.873\textsubscript{.015}} & \underline{.873\textsubscript{.004}} & \textbf{.921\textsubscript{.004}} \\
\midrule
\multirow{7}{*}{\rotatebox{90}{Qwen3.6}} & \cellcolor{gray!20}Source & \cellcolor{gray!20}.794 & \cellcolor{gray!20}.620 & \cellcolor{gray!20}.700 & \cellcolor{gray!20}.874 & \cellcolor{gray!20}.906 & \cellcolor{gray!20}.838 & \cellcolor{gray!20}.940 & \cellcolor{gray!20}.614 & \cellcolor{gray!20}.931 \\ \cmidrule{2-11}
 & Random & .759\textsubscript{.013} & .208\textsubscript{.152} & .317\textsubscript{.124} & .469\textsubscript{.062} & .625\textsubscript{.100} & .528\textsubscript{.115} & .580\textsubscript{.151} & .315\textsubscript{.226} & .604\textsubscript{.097} \\
 & Frequency & .789\textsubscript{.005} & \cellcolor{green!20}.681\textsubscript{.097} & .657\textsubscript{.005} & \underline{.812\textsubscript{.026}} & .848\textsubscript{.006} & .800\textsubscript{.029} & \textbf{.927\textsubscript{.021}} & \underline{\cellcolor{green!20}.787\textsubscript{.131}} & .907\textsubscript{.004} \\
 & Gate & .794\textsubscript{.002} & \underline{\cellcolor{green!20}.739\textsubscript{.019}} & .654\textsubscript{.007} & .788\textsubscript{.019} & \underline{.853\textsubscript{.008}} & \underline{.802\textsubscript{.020}} & .897\textsubscript{.015} & .517\textsubscript{.142} & .907\textsubscript{.011} \\
 & EAN & \underline{\cellcolor{green!20}.796\textsubscript{.007}} & .506\textsubscript{.196} & .319\textsubscript{.101} & .669\textsubscript{.015} & .745\textsubscript{.038} & .748\textsubscript{.019} & .783\textsubscript{.012} & \cellcolor{green!20}.751\textsubscript{.073} & .859\textsubscript{.004} \\
 & EASY-EP & .789\textsubscript{.001} & \textbf{\cellcolor{green!20}.822\textsubscript{.012}} & \textbf{.675\textsubscript{.002}} & \textbf{.835\textsubscript{.023}} & \textbf{.863\textsubscript{.011}} & \textbf{.813\textsubscript{.039}} & .903\textsubscript{.021} & \textbf{\cellcolor{green!20}.926\textsubscript{.006}} & \underline{.914\textsubscript{.004}} \\
 & REAP & \textbf{\cellcolor{green!20}.797\textsubscript{.002}} & .550\textsubscript{.057} & \underline{.663\textsubscript{.005}} & \textbf{.835\textsubscript{.015}} & \underline{.853\textsubscript{.010}} & .778\textsubscript{.038} & \underline{.910\textsubscript{.020}} & .271\textsubscript{.061} & \textbf{.921\textsubscript{.008}} \\
\bottomrule
\end{tabular}
}
\caption{Downstream performance on biomedical multiple-choice and QA datasets. Scores that are better than Source are highlighted in \colorbox{green!20}{green}. The best and second-best adaptation approaches for each model family are indicated in \textbf{bold} and \underline{underlined}, respectively. Subscripts denote standard deviation across three seeds.}
\label{tab:utility_cls_target}
\end{table*}